\documentclass[conference,a4paper]{IEEEtran}
\IEEEoverridecommandlockouts

\usepackage{iftex}
\ifXeTeX 
  \usepackage{fontspec}
\fi
\usepackage[a4paper,top=0.75in,bottom=1in,left=0.63in,right=0.63in,columnsep=0.25in,
  headsep=0in,headheight=0in,footskip=0.5in]{geometry}
\usepackage{cite}
\usepackage{amsmath,amssymb,amsfonts}
\usepackage{graphicx}
\usepackage{booktabs}
\usepackage{array}
\usepackage{xcolor}
\usepackage{url}
\usepackage[hidelinks]{hyperref}
\hypersetup{pdfauthor={Muhammad Rizwan, Muhammad Awais Khan, Muhammad Ahmad Kamran},
  pdfkeywords={small-object detection, thermal imaging, FLIR ADAS, YOLO},
  pdftitle={A Paired RGB-Thermal Benchmark and the MaxSPD-YOLO Detector for Small Road-User Detection}}

\graphicspath{{figures/}}
\newcommand{\mAP}{AP$_{50:95}$}

\begin{document}

\title{A Paired RGB--Thermal Benchmark and the MaxSPD-YOLO Detector for Small Road-User Detection}

\author{%
\fontsize{9}{11}\selectfont
\begin{tabular}{@{}*{3}{>{\centering\arraybackslash}p{0.31\textwidth}}@{}}
{\fontsize{11}{13}\selectfont Muhammad Rizwan}\newline
\textit{Artificial Intelligence Technology}\newline
\textit{Centre (AITeC)}\newline
\textit{National Centre for Physics (NCP)}\newline
Islamabad 44000, Pakistan\newline
rizwan00277@gmail.com
&
{\fontsize{11}{13}\selectfont Muhammad Awais Khan}\newline
\textit{Artificial Intelligence Technology}\newline
\textit{Centre (AITeC)}\newline
\textit{National Centre for Physics (NCP)}\newline
Islamabad 44000, Pakistan\newline
mawaiskhan0486@gmail.com
&
{\fontsize{11}{13}\selectfont Muhammad Ahmad Kamran}\newline
\textit{Artificial Intelligence Technology}\newline
\textit{Centre (AITeC)}\newline
\textit{National Centre for Physics (NCP)}\newline
Islamabad 44000, Pakistan\newline
malik.ahmad@ncp.edu.pk
\end{tabular}%
}

\maketitle
\pagestyle{empty}

\begin{abstract}
Detecting distant pedestrians, cyclists, and cars remains challenging for lightweight detectors, as these objects
often occupy only a few pixels in driving images. This study investigates whether architectural changes that
improve small-object detection can benefit both RGB and thermal modalities. We construct a paired benchmark from
FLIR ADAS v1.3, annotating 52,857 objects in RGB images across 9,598 paired frames with shared data splits.
Using this benchmark, we evaluate eleven YOLO configurations in both modalities with three training seeds each
(66 runs). The configuration selected on validation, MaxSPD-YOLO11n, combines higher-resolution (P2--P4)
detection heads, space-to-depth downsampling, and a MaxViT block. Compared with YOLO11n, it improves test
AP$_{50:95}$ from 37.9\% to 40.3\% in RGB and from 28.2\% to 29.9\% in thermal. Small-object AP also increases
from 31.5\% to 36.0\% in RGB and from 21.0\% to 22.8\% in thermal. The model has 43\% fewer parameters but
requires more computation and has a higher forward-pass latency (13.8 vs.\ 6.4 GFLOPs; 1.68 vs.\ 0.92\,ms). Ablation
results suggest that the P2--P4 detection layout accounts for much of the small-object improvement.
\end{abstract}


\section{Introduction}

A driver-assistance system must detect pedestrians, cyclists, and cars while they are still far away, because a
distant road user is one that the vehicle still has time to react to. At that distance the objects cover only a few
pixels, and lightweight detectors, which are the ones that fit on vehicle hardware, find them hardest.

Small objects are difficult because nano YOLO models predict at strides 8--32 and discard detail through repeated
downsampling, so a $12{\times}12$-pixel pedestrian covers fewer than two cells of the finest feature map. The two
common driving sensors provide different information. Visible-spectrum (RGB) images carry color and texture but
lose small objects at night and in glare; long-wave infrared (thermal) images show warm road users regardless of
illumination, with little texture and at lower resolution \cite{hwang2015kaist,jia2021llvip}. Whether a
detector change that helps small objects in one modality also helps in the other is therefore an open,
empirical question.

Answering it requires the \emph{same scenes} labeled in \emph{both} modalities. FLIR ADAS v1.3
\cite{flir_adas} is a widely used thermal driving dataset \cite{devaguptapu2019borrow,ghose2019saliency,
zhang2020cfr,ding2024tiryolo} that records synchronized RGB frames, but only the thermal frames are labeled. The
FLIR studies cited here train on thermal only \cite{ghose2019saliency,ding2024tiryolo}, use RGB without labels
\cite{devaguptapu2019borrow}, or transfer the thermal boxes to registered RGB crops \cite{zhang2020cfr,
guo2024damsdet}. Transferred boxes also mark objects that are invisible in RGB. A paired comparison needs RGB
boxes drawn in the RGB images, the same frames in both modalities, and identical splits. We do not fuse the
modalities: each detector sees one modality and is scored against that modality's own labels, so that the effect
of a design change can be examined in RGB and in thermal separately.

We build such a benchmark and use it for a controlled study of small road-user detection. Our contributions are:
\begin{enumerate}
  \item 52,857 RGB boxes for FLIR ADAS v1.3 on 9,598 frames, with split lists shared by both modalities, a
  train/validation/test similarity check, and day/night tags.
  \item An analysis of small-object detection in RGB and thermal images under modality-specific labels.
  \item MaxSPD-YOLO11n, a configuration of established components (P2--P4 layout, SPD, MaxViT), selected by a
  per-component ablation that also tests a P5 level and GAM attention.
  \item A comparison of eleven configurations, two modalities, and three seeds (66 runs) under one protocol, with
  model size, GFLOPs, and measured latency.
\end{enumerate}
Small objects are the weakest size bin of every detector, most of all in thermal images, and moving the detection
heads to P2--P4 contributes most of the small-object gain in both modalities, according to the ablation. 

 The rest of the paper covers related work (Section II), the dataset and annotation (Section III), the MaxSPD-YOLO configuration (Section IV), the experimental setup (Section V), results and discussion (Section VI), and conclusions (Section VII).

\section{Related Work}

KAIST \cite{hwang2015kaist} and LLVIP \cite{jia2021llvip} provide aligned
RGB--thermal pairs for pedestrians; FLIR-aligned
\cite{zhang2020cfr} registers a subset of FLIR and reuses the thermal boxes for RGB; recent multispectral
detectors such as DAMSDet \cite{guo2024damsdet} are evaluated on it. Fusion methods combine the two modalities
\cite{guo2024damsdet}. We study single-modality detectors on
labels drawn in each image, which fusion methods can build on.

YOLO necks merge scales with FPN/PAN \cite{lin2017fpn,
liu2018panet}, and a stride-4 (P2) prediction level is a common way to recover small objects, for example in
drone imagery \cite{zhu2021tph}. Recent thermal detectors adapt YOLO-type models to the FLIR ADAS thermal data
\cite{ding2024tiryolo} or improve the detection of small infrared objects with super-resolution and attention-based fusion
\cite{wang2025isryolo}.

SPD-Conv \cite{sunkara2022spd} replaces strided
convolutions with a lossless space-to-depth step, GAM \cite{liu2021gam} reweights channels and positions, and
MaxViT \cite{tu2022maxvit} adds global context at linear cost. Ultralytics YOLOv8 \cite{jocher2023yolov8}, YOLO11
\cite{jocher2024yolo11} and YOLO26 \cite{ultralytics2025yolo26} provide nano models of 2--3\,M parameters.
MaxSPD-YOLO combines these established components; our contribution is their controlled evaluation on paired
RGB and thermal data.

\section{Dataset and Annotation}
\label{sec:dataset}

FLIR v1.3 provides $640{\times}512$ thermal frames and higher-resolution RGB
frames that share a frame ID but are not pixel-aligned: the field of view and lens differ, so boxes cannot be
transferred between modalities. Three annotators labeled 9,617 of the 9,620 RGB frames of the FLIR train and
val splits (all have a thermal counterpart) in CVAT, with axis-aligned boxes for \emph{person}, \emph{bicycle},
\emph{car}, and \emph{dog}, the classes present in the thermal labels. Every object visible in the RGB image was
labeled, whether or not it is labeled in thermal; objects not visible in RGB were not labeled, even when FLIR
labels them in thermal. A reviewer then checked every frame and corrected boxes where
needed. The thermal labels are FLIR's own; the two label sets are kept separate.

An automatic check found no out-of-range, zero-area, invalid-class, or duplicate boxes.
We removed 19 annotated frames from both modalities: 18 blank RGB images and one test frame with a missed
annotation. The remaining 9,598 frames hold all 52,857 RGB boxes (Table~\ref{tab:dataset}). Of these frames, 81
have an empty RGB label file but five or more thermal boxes; all 81 are tagged as night, consistent with objects
that are not visible enough in RGB. As a cross-modal consistency check, RGB has on average 1.3 fewer persons and
1.0 fewer cars per frame than thermal.

\begin{table}[!t]
\centering
\caption{Paired benchmark: both modalities use the same frames in every split. Small $=$ COCO small
($<32^2$ px at 640-px width); Obj/img $=$ mean boxes per image. Dog (148/20/10 RGB, 201/18/13 thermal train/val/test boxes) is omitted for space.}
\label{tab:dataset}
\setlength{\tabcolsep}{2.4pt}
\resizebox{\columnwidth}{!}{%
\begin{tabular}{llrrrrrrr}
\toprule
Labels & Split & Images & Boxes & Person & Bicycle & Car & Small & Obj/img \\
\midrule
RGB & train & 7,515 & 38,572 & 8,717 & 2,374 & 27,333 & 14,776 & 5.13 \\
 & val & 829 & 4,568 & 1,019 & 342 & 3,187 & 1,992 & 5.51 \\
 & test & 1,254 & 9,717 & 4,735 & 370 & 4,602 & 3,465 & 7.75 \\
\midrule
Thermal & train & 7,515 & 59,322 & 19,928 & 3,430 & 35,763 & 34,887 & 7.89 \\
 & val & 829 & 5,802 & 1,816 & 376 & 3,592 & 3,275 & 7.00 \\
 & test & 1,254 & 10,830 & 5,404 & 420 & 4,993 & 4,972 & 8.64 \\
\bottomrule
\end{tabular}}
\end{table}

FLIR's \emph{val} split is our \textbf{test} split (1,254 frames). FLIR \emph{train} is divided
into \textbf{train} (7,515) and \textbf{validation} (829 frames) in blocks of 50 consecutive frame IDs (about
25\,s of driving), so that neighboring frames mostly stay in one split. Of 500 random block assignments, we keep
the one in which every class occurs in validation for both modalities and each class's validation share is
closest to 10\% (maximum deviation: 1.0 percentage point). Both modalities use the same frame lists.

We compare every frame with all frames of the other splits using the cosine similarity of
$40{\times}32$ grayscale RGB thumbnails. No test frame reaches 0.95 with any train or validation frame (maximum 0.91);
the closest pairs show the same street on a different drive. Validation is used only for checkpoint and design
selection, and the test split is evaluated once.

FLIR v1.3 has no time-of-day metadata, and whole-image brightness fails because RGB
auto-exposure makes lit night streets look bright. We threshold the mean HSV value of the top quarter of each RGB
image, which is mostly sky, with Otsu's method \cite{otsu1979}, and smooth the tag by a majority vote over
$\pm5$ neighboring frames. A visual check of 48 frames, including the 12 closest to the threshold, found no
errors.

\begin{figure}[!t]
  \centering
  \includegraphics[width=\columnwidth]{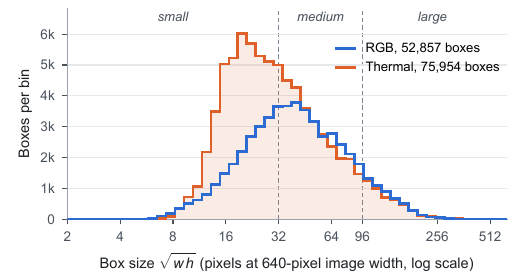}
  \caption{Box-size distribution of RGB (our labels) and thermal (FLIR labels) on the same 9,598 paired frames.
  Dashed lines mark the COCO size bins (32 and 96 pixels). Thermal labels contain many more small boxes.}
  \label{fig:sizes_dist}
\end{figure}

\begin{figure}[!t]
  \centering
  \includegraphics[width=0.8\columnwidth]{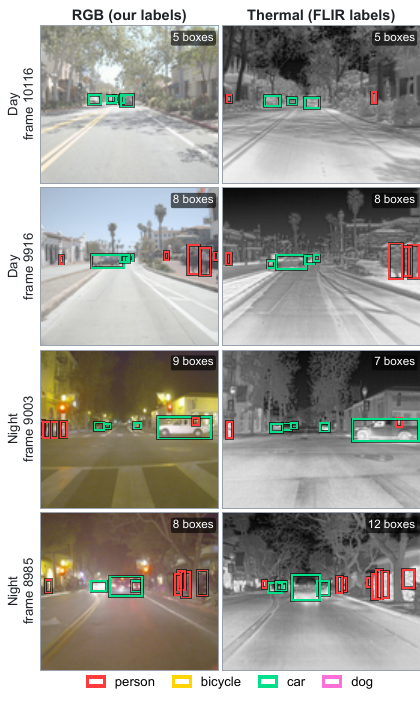}
  \caption{Paired day and night test frames with our RGB labels (left) and FLIR thermal labels (right), with the
  number of boxes per image. The cameras differ in field of view, so the images are not pixel-aligned. At night,
  several pedestrians labeled in thermal are not visible enough in RGB to be labeled there.}
  \label{fig:examples}
\end{figure}

On the same frames, the thermal labels contain 75,954 boxes against 52,857 in
RGB. The difference lies mainly in \emph{small} objects (COCO: ${<}32^2$ pixels at 640-pixel width): 34,887 small
thermal training boxes (59\% of thermal training boxes) against 14,776 in RGB (38\%; Fig.~\ref{fig:sizes_dist}),
and in night frames, with 22,402 thermal training boxes against 8,838 in RGB. Distant, warm pedestrians and cars
remain visible in thermal when they are too dark or too small to label in RGB (Fig.~\ref{fig:examples}). The
thermal task therefore contains many more small targets, and the two label sets describe the objects visible in
each modality.

\begin{figure*}[!t]
  \centering
  \includegraphics[width=\textwidth]{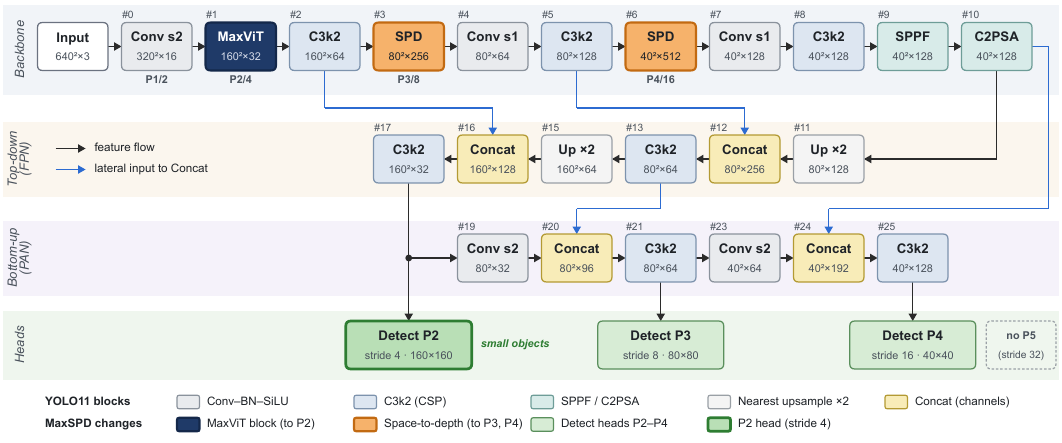}
  \caption{The proposed MaxSPD-YOLO11n. Compared with YOLO11n, a MaxViT block (dark blue) replaces the second
  stride-2 convolution and adds attention at stride 4; SPD (orange) followed by a stride-1 convolution replaces the
  strided convolutions to P3 and P4, so downsampling discards no pixels; and the heads move from P3--P5 to
  P2--P4, adding a stride-4 head for small objects and removing the P5 level. Layer numbers follow the model
  configuration; \#14, \#18, \#22 and \#26 are identity layers (GAM positions in MaxSPD-YOLO11n+GAM) and are not
  drawn. Shapes are $H^2{\times}C$ (height = width) for a $640{\times}640$ input.}
  \label{fig:arch}
\end{figure*}

\section{MaxSPD-YOLO}
\label{sec:method}

MaxSPD-YOLO (Fig.~\ref{fig:arch}) keeps the YOLO11n blocks and changes how spatial detail is carried to the
heads, because small objects lose most of their information in early downsampling and at coarse prediction
strides.

The FPN/PAN neck is shifted one level toward higher resolution (P2 instead of P5). The
top-down (FPN) path upsamples deep features and concatenates them with backbone features at P3 and P2; the
bottom-up (PAN) path downsamples again and concatenates with the top-down P3 features and the backbone P4
output. Three decoupled Detect heads predict at strides 4 ($160{\times}160$), 8 and 16; at stride 4 a
$12{\times}12$-pixel pedestrian covers $3{\times}3$ cells. Removing the P5 level and its wide channels pays for
part of the high-resolution path.

A stride-2 convolution samples one position in four. SPD \cite{sunkara2022spd}
moves every $2{\times}2$ neighborhood into channels,
\begin{equation}
  \mathrm{SPD}(X) = [X_{0::2,0::2} \,\|\, X_{1::2,0::2} \,\|\, X_{0::2,1::2} \,\|\, X_{1::2,1::2}],
\end{equation}
mapping $H{\times}W{\times}C$ to $\frac{H}{2}{\times}\frac{W}{2}{\times}4C$ without discarding pixels; a
$3{\times}3$ stride-1 convolution then reduces the channels. SPD replaces the backbone's stride-2 convolutions to
P3 and P4; the two stride-2 convolutions of the PAN path are unchanged.

A MaxViT block \cite{tu2022maxvit} replaces the stride-2 convolution to P2. It consists of
an MBConv layer (expansion 4, depthwise $3{\times}3$ convolution with stride 2, squeeze-and-excitation) followed by block attention in non-overlapping $8{\times}8$ windows and grid attention over a
dilated $8{\times}8$ grid. Its cost is linear in the number of pixels, so it can add global context at stride 4.

GAM \cite{liu2021gam} can follow each of the four neck C3k2 blocks; it applies channel
attention with a per-pixel MLP (ratio 4) and spatial attention with two $7{\times}7$ convolutions.

The proposed model, \textbf{MaxSPD-YOLO11n}, uses the P2--P4 layout, SPD and MaxViT without
GAM; it is ablation step A3 and was selected on validation (Section~\ref{sec:ablation}). MaxSPD-YOLO11n+GAM and
MaxSPD-YOLOv8n+GAM add GAM after each neck C3k2 block of the same layout, built on YOLO11n and YOLOv8n blocks, respectively. MaxSPD-YOLO26n-S is a slim YOLO26n
version ($1{\times}1$ convolutions after SPD, no MBConv expansion, depthwise GAM) that stays within 10\% of the
YOLO26n GFLOPs. MaxSPD-YOLO11n has 1.47\,M parameters, 43\% fewer than YOLO11n.

\section{Experimental Setup}

\begin{table*}[!t]
\centering
\caption{Main results on the test split (FLIR official \emph{val}, 1,254 paired frames). COCO AP (\%) from
pycocotools, mean $\pm$ std over three seeds; all models trained from scratch with the same protocol. Params and
GFLOPs are for the fused model at $640{\times}640$; latency is the CUDA-graph forward pass at batch 1, FP16, on an
RTX~4070~Ti~SUPER, without preprocessing or NMS. Best accuracy per column in bold. YOLO11s is a larger reference model.
MaxSPD-YOLO11n (A3) is the proposed model, selected on the validation split; +GAM marks variants with GAM attention.}
\label{tab:main}
\begin{tabular}{lrrrcccc}
\toprule
 & & & & \multicolumn{2}{c}{RGB (our labels)} & \multicolumn{2}{c}{Thermal (FLIR labels)} \\
\cmidrule(lr){5-6}\cmidrule(lr){7-8}
Model & Params (M) & GFLOPs & Latency (ms) & AP$_{50}$ & AP$_{50:95}$ & AP$_{50}$ & AP$_{50:95}$ \\
\midrule
YOLOv8n & 3.01 & 8.1 & 0.82 & 59.5{\scriptsize$\pm$2.2} & 39.3{\scriptsize$\pm$0.8} & 54.6{\scriptsize$\pm$0.6} & 28.5{\scriptsize$\pm$0.3} \\
YOLO11n & 2.58 & 6.4 & 0.92 & 56.9{\scriptsize$\pm$0.3} & 37.9{\scriptsize$\pm$0.3} & 53.3{\scriptsize$\pm$0.1} & 28.2{\scriptsize$\pm$0.3} \\
YOLO26n & 2.38 & 5.3 & 0.99 & 57.2{\scriptsize$\pm$0.5} & 37.8{\scriptsize$\pm$0.3} & 53.3{\scriptsize$\pm$0.1} & 28.1{\scriptsize$\pm$0.4} \\
\midrule
YOLO11s & 9.41 & 21.4 & 1.35 & \textbf{59.7}{\scriptsize$\pm$1.1} & 39.6{\scriptsize$\pm$0.6} & \textbf{56.5}{\scriptsize$\pm$0.8} & 30.0{\scriptsize$\pm$0.4} \\
\midrule
MaxSPD-YOLOv8n+GAM & 1.93 & 18.8 & 1.89 & 59.4{\scriptsize$\pm$0.6} & 40.2{\scriptsize$\pm$0.8} & 55.4{\scriptsize$\pm$0.5} & 29.9{\scriptsize$\pm$0.2} \\
MaxSPD-YOLO26n-S & 0.57 & 5.7 & 1.56 & 57.3{\scriptsize$\pm$0.4} & 37.8{\scriptsize$\pm$0.5} & 52.7{\scriptsize$\pm$0.1} & 28.0{\scriptsize$\pm$0.1} \\
MaxSPD-YOLO11n+GAM & 2.11 & 19.1 & 2.02 & 59.3{\scriptsize$\pm$0.3} & \textbf{40.3}{\scriptsize$\pm$0.2} & 55.5{\scriptsize$\pm$0.3} & \textbf{30.1}{\scriptsize$\pm$0.1} \\
MaxSPD-YOLO11n (A3) & 1.47 & 13.8 & 1.68 & 59.4{\scriptsize$\pm$0.4} & \textbf{40.3}{\scriptsize$\pm$0.2} & 56.0{\scriptsize$\pm$0.5} & 29.9{\scriptsize$\pm$0.1} \\
\bottomrule
\end{tabular}
\end{table*}

The baselines are YOLOv8n, YOLO11n, and YOLO26n, plus the larger YOLO11s (9.41\,M parameters,
21.4 GFLOPs) as a reference. MaxSPD variants are built on YOLOv8n, YOLO11n, and YOLO26n; the ablation adds four
YOLO11n-based variants (A1--A4), one of which (A3) is MaxSPD-YOLO11n. Each of the eleven configurations is
trained on RGB and on thermal with seeds 0, 1, and 2 (66 runs).

All models are trained from scratch, because the MaxSPD layers have no pretrained weights:
$640{\times}640$ input, 150 epochs without early stopping, SGD (momentum 0.937, weight decay $5{\cdot}10^{-4}$),
learning rate $0.01\rightarrow10^{-4}$ (linear decay, 3 warm-up epochs), Ultralytics default augmentation with
mosaic off for the last 10 epochs, and mixed precision. The batch size is 16 for MaxSPD-YOLOv8n+GAM,
MaxSPD-YOLO11n+GAM and A1--A4, which did not fit into the 16\,GB GPU memory at batch size 32, and 32 for the other models; gradients are
accumulated to a nominal batch of 64 in all cases. To keep one protocol for all models, YOLO26n and
MaxSPD-YOLO26n-S are also trained with SGD rather than the MuSGD optimizer that Ultralytics' automatic setting
selects for runs of this length, and they are evaluated, like all other models, with the one-to-many head and NMS
(the Ultralytics 8.4.173 default), not with the NMS-free one-to-one head. The checkpoint with the best validation
fitness is kept, and each frozen model is evaluated once on the test split.

We report COCO AP \cite{lin2014coco} computed with \texttt{pycocotools} (confidence 0.001, NMS IoU 0.7, at most
300 detections): AP$_{50}$, \mAP{}, AP$_S$/AP$_M$/AP$_L$ (COCO size bins at 640-pixel width), and \mAP{} on
day-only and night-only frames, as mean $\pm$ std over three seeds. Latency is measured for the network forward pass of the
fused model, replayed as a CUDA graph on one NVIDIA RTX~4070~Ti~SUPER (batch 1, FP16, median of three interleaved
rounds of 500 passes); preprocessing and NMS are excluded. GFLOPs count arithmetic operations, whereas latency
depends on the implementation and hardware, so we report both. Software: PyTorch 2.11 (CUDA 12.8), Ultralytics
8.4.173.

We compare models with Welch's $t$-test on the three per-seed test \mAP{} values of each
model ($n{=}3$ vs.\ 3) and also report whether every seed of one model scores above every seed of the other. These
tests reflect variation between training runs, not between test images, and a small $p$-value does not by itself
indicate a large practical difference.

\section{Results and Discussion}
\label{sec:results}

\subsection{Main Comparison}

Among the stock nano models, YOLOv8n is the most accurate (Table~\ref{tab:main}). MaxSPD-YOLO11n scores
higher than YOLO11n: \mAP{} rises from 37.9 to 40.3 on RGB ($+$2.4, $p{=}0.0004$) and from 28.2 to 29.9 on thermal ($+$1.7,
$p{=}0.005$), and each of its seeds scores above each YOLO11n seed in both modalities. The same layout with GAM
also raises the scores of YOLOv8n by $+$0.9 (RGB) and $+$1.4 (thermal, $p{=}0.009$, every seed higher). Against the larger
YOLO11s, MaxSPD-YOLO11n is on par in overall \mAP{} ($+$0.7 RGB, $p{=}0.16$; $-$0.1 thermal, $p{=}0.72$) with
6.4 times fewer parameters and fewer GFLOPs (13.8 vs.\ 21.4), although its measured latency is higher (1.68 vs.\\
1.35\,ms), and it exceeds YOLO11s on small RGB objects (Section~\ref{sec:sizes}).

\begin{table*}[!t]
\centering
\caption{Ablation on the YOLO11n backbone, adding one MaxSPD component at a time (removed modules are replaced by
identity layers, so all other layers are unchanged; A1 uses stride-2 convolutions where SPD would be).
AP$_{50:95}$ and small-object AP$_S$ (\%), mean $\pm$ std over three seeds; ms = FP16 batch-1 GPU forward latency as in
Table~\ref{tab:main}. A3 is MaxSPD-YOLO11n and Full is MaxSPD-YOLO11n+GAM. Design decisions use the validation columns only.}
\label{tab:ablation}
\setlength{\tabcolsep}{3.2pt}
\begin{tabular}{lrrrcccccccc}
\toprule
 & & & & \multicolumn{4}{c}{AP$_{50:95}$} & \multicolumn{4}{c}{AP$_S$ (small objects)} \\
\cmidrule(lr){5-8}\cmidrule(lr){9-12}
 & & & & \multicolumn{2}{c}{Validation} & \multicolumn{2}{c}{Test} & \multicolumn{2}{c}{Validation}
 & \multicolumn{2}{c}{Test} \\
\cmidrule(lr){5-6}\cmidrule(lr){7-8}\cmidrule(lr){9-10}\cmidrule(lr){11-12}
Variant & Params (M) & GFLOPs & ms & RGB & Thermal & RGB & Thermal & RGB & Thermal & RGB & Thermal \\
\midrule
YOLO11n (P3--P5) & 2.58 & 6.4 & 0.92 & 32.7{\scriptsize$\pm$0.3} & 32.5{\scriptsize$\pm$1.1} & 37.9{\scriptsize$\pm$0.3} & 28.2{\scriptsize$\pm$0.3} & 26.8{\scriptsize$\pm$0.6} & 24.9{\scriptsize$\pm$0.9} & 31.5{\scriptsize$\pm$0.7} & 21.0{\scriptsize$\pm$0.7} \\
A1: P2--P4 heads & 0.88 & 9.3 & 0.99 & 34.7{\scriptsize$\pm$0.3} & 34.8{\scriptsize$\pm$0.8} & 39.0{\scriptsize$\pm$0.9} & 28.8{\scriptsize$\pm$0.2} & 30.0{\scriptsize$\pm$0.7} & 27.7{\scriptsize$\pm$0.5} & 35.9{\scriptsize$\pm$3.2} & 22.2{\scriptsize$\pm$1.4} \\
A2: A1 + SPD & 1.43 & 12.1 & 1.04 & 33.8{\scriptsize$\pm$0.6} & 34.1{\scriptsize$\pm$1.2} & 38.5{\scriptsize$\pm$0.9} & 29.2{\scriptsize$\pm$0.3} & 29.1{\scriptsize$\pm$0.3} & 27.2{\scriptsize$\pm$0.8} & 33.7{\scriptsize$\pm$1.3} & 21.9{\scriptsize$\pm$0.3} \\
A3: A2 + MaxViT & 1.47 & 13.8 & 1.68 & \textbf{35.6}{\scriptsize$\pm$0.9} & \textbf{35.8}{\scriptsize$\pm$0.7} & 40.3{\scriptsize$\pm$0.2} & 29.9{\scriptsize$\pm$0.1} & 30.3{\scriptsize$\pm$0.8} & \textbf{28.1}{\scriptsize$\pm$0.5} & 36.0{\scriptsize$\pm$0.4} & 22.8{\scriptsize$\pm$0.2} \\
A4: A3 + P5 head & 4.13 & 15.5 & 1.94 & 34.3{\scriptsize$\pm$0.4} & 34.5{\scriptsize$\pm$0.7} & \textbf{40.6}{\scriptsize$\pm$1.0} & 29.6{\scriptsize$\pm$0.6} & 29.9{\scriptsize$\pm$0.8} & 27.5{\scriptsize$\pm$0.7} & \textbf{36.4}{\scriptsize$\pm$1.0} & \textbf{23.5}{\scriptsize$\pm$0.5} \\
Full: A3 + GAM & 2.11 & 19.1 & 2.02 & 34.7{\scriptsize$\pm$0.4} & 34.9{\scriptsize$\pm$0.7} & 40.3{\scriptsize$\pm$0.2} & \textbf{30.1}{\scriptsize$\pm$0.1} & \textbf{30.4}{\scriptsize$\pm$0.4} & 28.0{\scriptsize$\pm$0.3} & 36.0{\scriptsize$\pm$0.4} & 22.8{\scriptsize$\pm$0.4} \\
\bottomrule
\end{tabular}
\end{table*}

\subsection{Ablation}
\label{sec:ablation}

Table~\ref{tab:ablation} adds the components to YOLO11n one at a time; design choices use the validation
columns. \textbf{A1} keeps the YOLO11n blocks and moves the heads to P2--P4. With 0.88\,M parameters it gains
2.0 (RGB) and 2.3 (thermal) validation \mAP{} and 3.2 and 2.8 validation AP$_S$, about 90\% of the validation
AP$_S$ gain of A3 in both modalities. \textbf{A2} adds SPD and \textbf{A3} adds MaxViT; A3 is 1.7--1.8 \mAP{}
and 0.9--1.2 AP$_S$ above A2 on validation and has the best validation \mAP{} in both modalities. \textbf{A4} adds
a P5 level and head back to A3, at 2.7\,M extra parameters; it is 1.3 points below A3 in validation \mAP{} in both
modalities. Adding \textbf{GAM} to A3 costs 0.64\,M parameters, 5.2 GFLOPs (5.3 from the rounded values in
Table~\ref{tab:ablation}) and 0.34\,ms, and gives 0.9 points less validation \mAP{} in both modalities. Based on the
validation results, we therefore select A3, without P5 and GAM, as MaxSPD-YOLO11n.

\begin{table}[!t]
\centering
\caption{Test AP$_{50:95}$ (\%) by object size (COCO bins) and by time of day, mean $\pm$ std over three seeds.}
\label{tab:breakdown}
\setlength{\tabcolsep}{2.2pt}
\resizebox{\columnwidth}{!}{%
\begin{tabular}{lccccc}
\toprule
Model & AP$_S$ & AP$_M$ & AP$_L$ & Day & Night \\
\midrule
\multicolumn{6}{l}{\emph{RGB}} \\
\quad YOLO11n & 31.5{\scriptsize$\pm$0.7} & 41.0{\scriptsize$\pm$0.9} & 58.9{\scriptsize$\pm$2.2} & 39.7{\scriptsize$\pm$0.4} & 44.4{\scriptsize$\pm$0.4} \\
\quad YOLO26n & 31.9{\scriptsize$\pm$0.9} & 41.4{\scriptsize$\pm$1.6} & 57.7{\scriptsize$\pm$1.8} & 39.6{\scriptsize$\pm$0.2} & 43.4{\scriptsize$\pm$0.1} \\
\quad YOLOv8n & 34.4{\scriptsize$\pm$2.3} & 41.9{\scriptsize$\pm$1.6} & 61.4{\scriptsize$\pm$2.9} & 41.3{\scriptsize$\pm$0.8} & 44.3{\scriptsize$\pm$1.0} \\
\quad YOLO11s & 34.7{\scriptsize$\pm$0.6} & 42.1{\scriptsize$\pm$1.1} & \textbf{63.3}{\scriptsize$\pm$1.7} & 41.7{\scriptsize$\pm$0.4} & 45.1{\scriptsize$\pm$1.3} \\
\quad MaxSPD-YOLO11n+GAM & \textbf{36.0}{\scriptsize$\pm$0.4} & 42.8{\scriptsize$\pm$0.1} & 56.2{\scriptsize$\pm$1.0} & \textbf{42.2}{\scriptsize$\pm$0.4} & \textbf{47.1}{\scriptsize$\pm$0.8} \\
\quad MaxSPD-YOLO11n (A3) & \textbf{36.0}{\scriptsize$\pm$0.4} & \textbf{43.2}{\scriptsize$\pm$0.5} & 55.3{\scriptsize$\pm$0.9} & 42.1{\scriptsize$\pm$0.4} & 46.9{\scriptsize$\pm$0.9} \\
\midrule
\multicolumn{6}{l}{\emph{Thermal}} \\
\quad YOLO11n & 21.0{\scriptsize$\pm$0.7} & 32.6{\scriptsize$\pm$0.1} & 46.4{\scriptsize$\pm$2.4} & 27.1{\scriptsize$\pm$0.4} & 40.2{\scriptsize$\pm$1.1} \\
\quad YOLO26n & 23.0{\scriptsize$\pm$3.6} & 32.4{\scriptsize$\pm$0.5} & 47.9{\scriptsize$\pm$0.4} & 27.1{\scriptsize$\pm$0.3} & 39.7{\scriptsize$\pm$0.4} \\
\quad YOLOv8n & 21.3{\scriptsize$\pm$0.1} & 32.8{\scriptsize$\pm$0.3} & \textbf{49.3}{\scriptsize$\pm$1.5} & 27.2{\scriptsize$\pm$0.4} & 41.1{\scriptsize$\pm$0.2} \\
\quad YOLO11s & \textbf{23.1}{\scriptsize$\pm$0.9} & 34.3{\scriptsize$\pm$0.4} & 48.1{\scriptsize$\pm$2.0} & \textbf{28.9}{\scriptsize$\pm$0.6} & 42.6{\scriptsize$\pm$0.3} \\
\quad MaxSPD-YOLO11n+GAM & 22.8{\scriptsize$\pm$0.4} & \textbf{34.9}{\scriptsize$\pm$0.3} & 47.5{\scriptsize$\pm$1.8} & \textbf{28.9}{\scriptsize$\pm$0.2} & \textbf{43.0}{\scriptsize$\pm$0.6} \\
\quad MaxSPD-YOLO11n (A3) & 22.8{\scriptsize$\pm$0.2} & 34.6{\scriptsize$\pm$0.2} & 46.1{\scriptsize$\pm$1.1} & 28.7{\scriptsize$\pm$0.3} & 42.8{\scriptsize$\pm$0.6} \\
\bottomrule
\end{tabular}}
\end{table}

\begin{figure}[!t]
  \centering
  \includegraphics[width=\columnwidth]{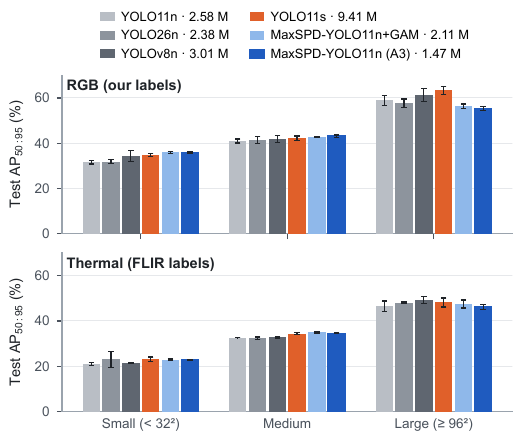}
  \caption{Test AP$_{50:95}$ by object size (COCO bins) for the six models of Table~\ref{tab:breakdown}, mean
  $\pm$ std over three seeds. Legend values are parameter counts (Table~\ref{tab:main}); MaxSPD-YOLO11n (A3) is
  the proposed model.}
  \label{fig:sizes_ap}
\end{figure}

\subsection{Object Size and Time of Day}
\label{sec:sizes}

Small objects are the weakest size bin of every detector in both modalities (Table~\ref{tab:breakdown},
Fig.~\ref{fig:sizes_ap}): YOLO11n
reaches a test AP$_S$ of 31.5 (RGB) and 21.0 (thermal) against an AP$_L$ of 58.9 and 46.4, so small thermal
objects are the hardest case. MaxSPD-YOLO11n raises AP$_S$ over YOLO11n on every seed, in both modalities and on
both splits: on test by $+$4.5 on RGB ($p{=}0.002$) and $+$1.8 on thermal ($p{=}0.04$), on validation by $+$3.5
and $+$3.2. On RGB its test AP$_S$ also exceeds that of YOLO11s (36.0 vs.\ 34.7). On thermal, medium objects gain
about as much (AP$_M$ $+$2.0), and on validation MaxSPD-YOLO11n has a higher thermal AP$_S$ than YOLO11s and
YOLO26n (28.1 vs.\ 27.1 and 26.1; not tabulated). Large objects do not follow this trend: AP$_L$ is lower than
for YOLO11n on RGB (55.3 vs.\ 58.9) and similar on thermal (46.1 vs.\ 46.4). The overall gain over YOLO11n holds on
both day and night frames in both modalities.

AP is measured against modality-specific labels, so cross-modal differences
describe the two benchmarks rather than rank the sensors. Every detector scores higher on RGB than on thermal test
frames (e.g., 37.9 vs.\ 28.2 for YOLO11n), and the thermal test set contains 4,972 small boxes against 3,465 in RGB,
many of them objects that are visible only in thermal. Both modalities also score higher at night than by day.
For RGB, this is likely a consequence of the labeling policy (only visible, usually nearer, and lit objects are labeled at
night: 5.7 boxes per night test image in RGB vs.\ 8.8 in thermal); for thermal, warm objects may contrast more
strongly with a cooler background. Validation and test AP also move in opposite directions (YOLO11n:
32.7$\rightarrow$37.9 RGB, 32.5$\rightarrow$28.2 thermal) because the test split, from separate FLIR drives, has
more night frames (42\% vs.\ 28\%) and more persons (49--50\% of boxes vs.\ 22--31\%). We therefore compare models
only within one split. Bicycles are the most difficult frequent class in thermal (15.6 \mAP{} for YOLO11n) and gain the most
from MaxSPD-YOLO11n ($+$3.8).

\subsection{Efficiency}

Parameter count and computation must be distinguished. MaxSPD-YOLO11n has fewer parameters than every
baseline (1.47\,M vs.\ 2.38--9.41\,M; Table~\ref{tab:main}), but its stride-4 path processes $160{\times}160$ maps,
so it needs 13.8 GFLOPs, compared with 6.4 for YOLO11n. The
ablation separates the latency of each component (Table~\ref{tab:ablation}): the P2--P4 layout and SPD add little
(A1/A2: 0.99/1.04\,ms forward pass), MaxViT adds 0.64\,ms at almost no parameter cost, probably because its attention runs
on $160{\times}160$ maps, and GAM would add another 0.34\,ms. The slim YOLO26n variant has roughly the same GFLOPs as
YOLO26n but a higher latency, which suggests that GFLOPs alone do not predict latency for attention and
memory-bound layers.

The small-object gains come with higher computation (2.2$\times$ the GFLOPs and
1.8$\times$ the latency of YOLO11n) and a modest drop in large-object AP on RGB. 

\section{Conclusion}
This study investigated small road-user detection in RGB and thermal images using a paired benchmark of 9,598
frames from FLIR ADAS v1.3. The experiments suggest that architectural changes aimed at preserving spatial
information can improve small-object detection in both modalities. Among the evaluated configurations,
MaxSPD-YOLO11n achieved higher test AP$_{50:95}$ than YOLO11n, increasing from 37.9\% to 40.3\% in RGB and from
28.2\% to 29.9\% in thermal, and higher small-object AP (from 31.5\% to 36.0\% and from 21.0\% to 22.8\%).
Ablation results suggest that the higher-resolution P2--P4 detection layout accounts for much of this
improvement.

The RGB annotations (YOLO format, keyed by FLIR frame ID), split lists, day/night
tags, model configurations, and the training, evaluation, and tag-generation code will be released
publicly. FLIR images must be obtained from Teledyne FLIR \cite{flir_adas}.

\section*{Acknowledgment}
Claude (Anthropic) was used for language editing, proofreading, \LaTeX{} formatting, and assistance with code
implementation. All AI-assisted outputs were reviewed and verified by the authors, who take full responsibility for
the manuscript, methodology, and results.

\bibliographystyle{IEEEtran}
\bibliography{references}

\end{document}